\documentclass{article}
\usepackage{amsmath,amsfonts}
\usepackage{algorithmic}
\usepackage{algorithm}
\usepackage{array}
\usepackage[caption=false,font=normalsize,labelfont=sf,textfont=sf]{subfig}
\usepackage{textcomp}
\usepackage{stfloats}
\usepackage{url}
\usepackage{verbatim}
\usepackage{graphicx}
\usepackage{cite}
\usepackage{gensymb}

\newcommand{\msym}{S}

\begin{document}

\title{Alternative metrics to select motors for Quasi-Direct Drive actuators}

\author{Karthik Urs,
Challen Enninful Adu, Elliott J. Rouse,
Talia Y. Moore,
\thanks{K. Urs and C. Enninful Adu are with the EMBIR Lab, Robotics,
University of Michigan, Ann Arbor, MI 48109 USA email: ursk, enninful @umich.edu.}
\thanks{E. Rouse is with the Neurobionics Lab, Mechanical Engineering, Robotics,
University of Michigan, Ann Arbor, MI 48109 USA email: ejrouse@umich.edu.}
\thanks{T.Y. Moore is with the EMBIR Lab, Mechanical Engineering, Robotics, 
Ecology and Evolutionary Biology, and Museum of Zoology, University of Michigan, Ann Arbor, MI 48109 USA. email: taliaym@umich.edu}}

\maketitle

\begin{abstract}
Robotic systems for legged locomotion---including legged robots, exoskeletons, and prosthetics---require actuators with low inertia and high output torque.
Traditionally, motors have been selected for these applications by maximizing the motor gap radius.
We present alternative metrics for motor selection that are invariant to transmission ratio.
The proposed metrics reward minimizing the motor inertia while maximizing the torque and motor constants without special consideration for gap radius, providing a better balance of properties for legged locomotion applications.
We rigorously characterize the T-Motor RI50 and demonstrate the use of the metrics by comparing the RI50 to the widely-used T-Motor U8 as a case study.
\end{abstract}

\section{Introduction}
{Quasi-Direct} Drive Actuators (QDDs) have enabled high-speed and robust locomotion in legged robots
by combining a powerful motor with a low-ratio transmission to minimize output mechanical impedance (OMI) and mass while meeting torque and power requirements.
Designers must carefully balance competing performance properties when selecting motors to make successful QDDs for legged locomotion.


\begin{figure}[t]
    \centering
    \includegraphics[width=\columnwidth]{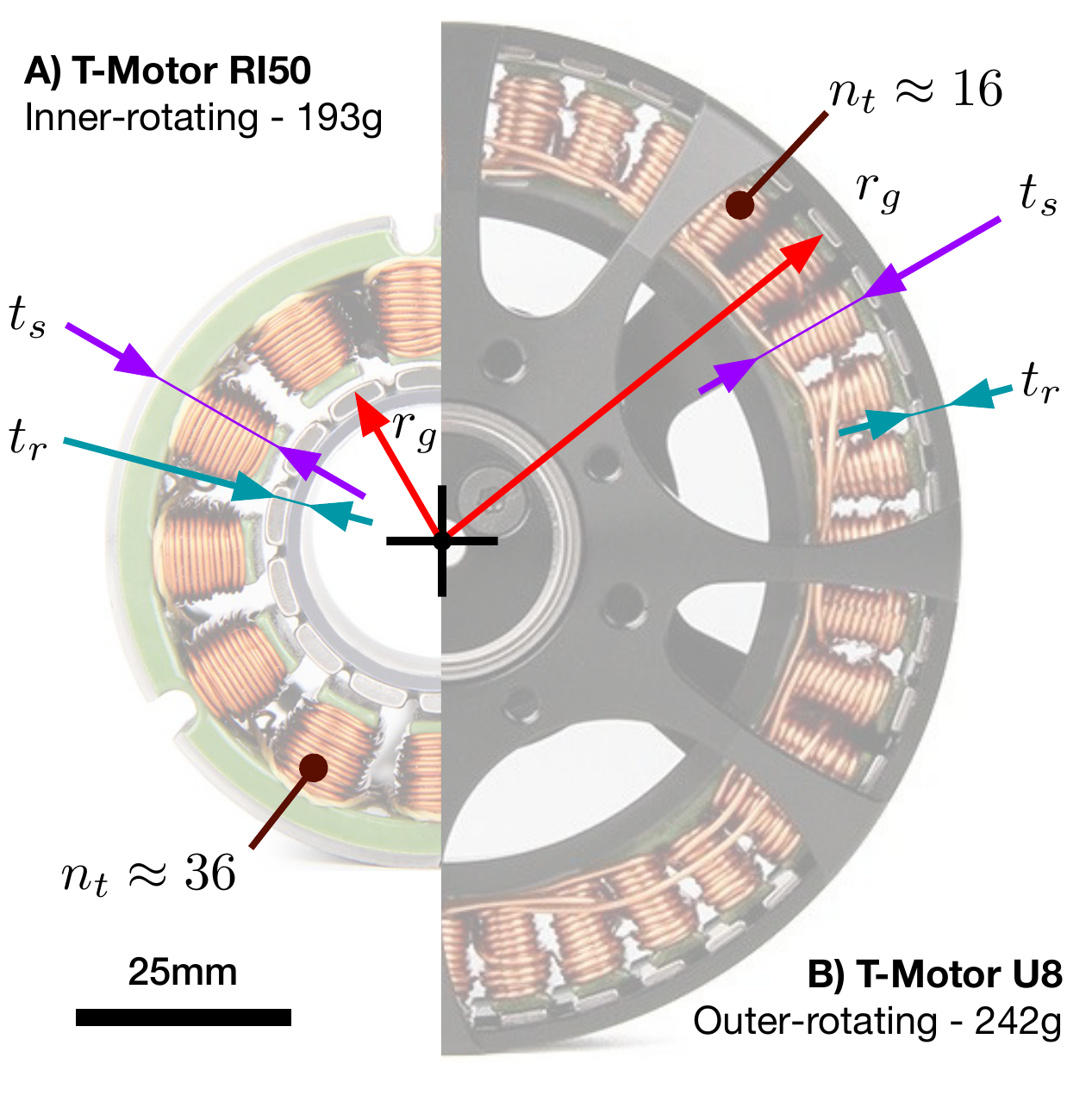}
    \caption{Physical characteristics of A) the RI50 and B) the U8 motors.}
    \label{fig:motorpics}
\end{figure}

Legged robotic systems apply large ground reaction forces (GRFs) to execute demanding actions, such as high-speed galloping, rapid direction changes, and jumps.
These actions impose large torque requirements for actuators. Because actuators make up a significant portion of total robot mass, they should be as lightweight as possible to translate actuator torques into high robot accelerations. 

Simultaneously, actuators should have low OMI, which is largely dominated by the rotational inertia of the actuator.
Low OMI, combined with modern commutation \cite{mevey2009sensorless}, allows the actuators to accurately output torques and act as torque sensors via measurement of motor current, providing valuable data to higher-level robot controllers.
This capability is analogous to neurological proprioception, so QDDs are also referred to proprioceptive actuators \cite{impactmitigation}.
With high OMI, every step of a legged robot 
generates high reaction forces in the robot structure, potentially resulting in damage.
Robot leg actions driven by the environment must overcome the OMI before acting on the motor, so high OMI also reduces proprioceptive accuracy (\textit{i.e.}, lower ``transparency'').


Thus, motors with a high torque constant $K_T$ (torque per current, [Nm/A]), high motor constant $K_M$ (torque per power heat waste, [Nm/$\sqrt{\text{W}}$]), low inertia $J_m$ [kgm$^2$], and low mass $m$ [kg] are desired.
Because these parameters are subject to tradeoffs, designers typically select a motor with an advantageous balance between key parameters above some minimum torque or power capability, and 
select a transmission to make best use of the motor for the given application (\textit{i.e.}, sufficiently amplify torque while retaining speed).

Popular design principles suggest that maximizing gap radius ($r_g$) of a motor is the best way to address the challenge of motor selection for QDDs; this strategy is founded on analysis of ideal motor models, trends shown on a large range of motors \cite{impactmitigation, seok_actuator_2012, zhu2021design, hanselman2003brushless}, and encouraging performance of modern robots with high-radius motors \cite{minicheetah, minitaur, azocar2020design}. However, for a specific application, the best motor choice may not have the highest $r_g$.
We propose a pair of simple selection metrics independent of $r_g$ to identify motors with 
useful properties for legged robots (Section \ref{sec:selection}).
We also provide a case study of 
the RI50 motor, 
which defies the larger-scale trends (Section \ref{sec:casestudy}).
Methods for characterizing the RI50 motor are described in Section \ref{sec:characterization}; 
results and comparisons are summarized in Section \ref{sec:results}.
A physics-based rationale to explain the observed differences between real motors and scaling-based predictions is proposed in Section \ref{sec:discussion}.

\section{Current Design Principles}
\label{sec:principles}

\begin{figure}[t]
    \centering
    \includegraphics[width=.9\columnwidth]{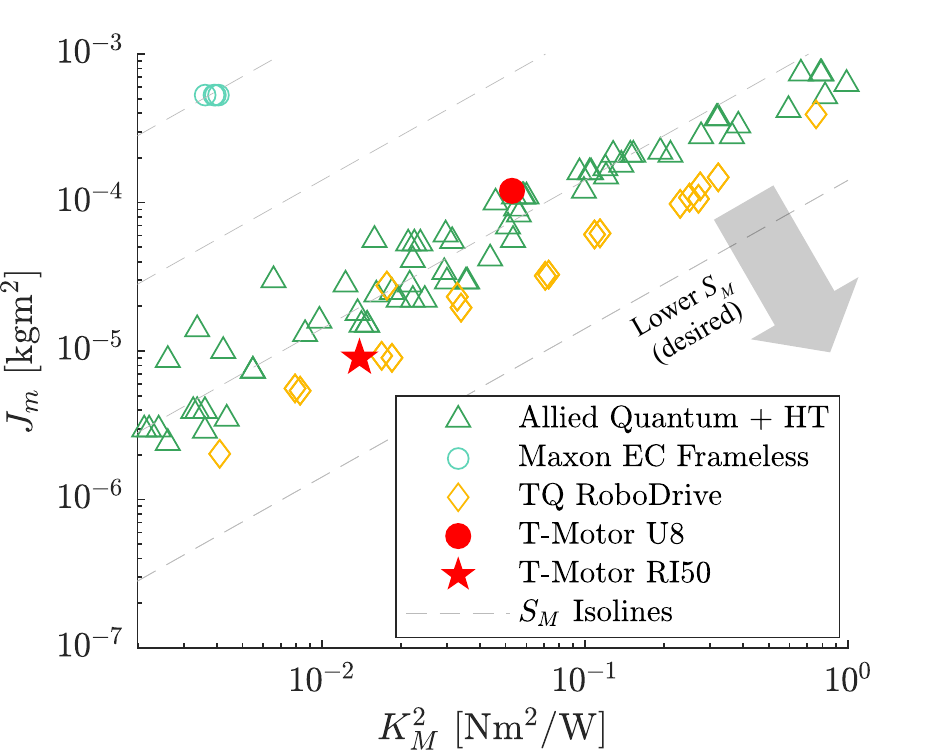}
    \caption{Commercial motor options and isolines with order-of-magnitude spacing for $\msym_M$ (\ref{eq:scaling_J_KM}). Motors closer to the bottom-right (such as the RI50) are considered more suitable for QDDs under the proposed metrics (lower $\msym_M$). U8 and RI50 empirically characterized; others from \cite{noauthor_maxons_nodate, noauthor_brushless_nodate, noauthor_robodrive_nodate}.}
    \label{fig:ideal}
\end{figure}

Several metrics and principles have been proposed to guide the motor selection step of QDD actuator design. 
Because temperature is frequently the limiting factor on continuous motor torque and power, 
motors can be selected by maximizing thermal specific torque, $K_\text{ts}$ [Nm/(kg$\sqrt{\degree \text{C}}$)], proposed in \cite{minitaur}:
\begin{equation}
    K_\text{ts} = \frac{K_T}{m}\sqrt{\frac{1}{R_\text{th} R_{\phi}}}
    \label{eq:therm_specific_torque}
\end{equation}
where $R_\text{th}$ is the motor thermal resistance in [$\degree$C/W] and $R_{\phi}$ is the electrical phase resistance of the motor [$\Omega$].

Others use more general principles derived from physics-based dimensional scaling analysis of ideal motor models.
Under a mass constraint and some given torque requirement, maximizing motor gap radius $r_g$ is often considered the best way to address the design priorities established above  \cite{impactmitigation, seok_actuator_2012, zhu2021design, hanselman2003brushless}. 
The following are developed in \cite{impactmitigation}, reproduced with minor nomenclature adjustments:
\begin{align}
    K_T &\propto r_g^2 l
    \label{eq:mit_kt} \\
    K_T/m &\propto r_g \\
    K_T/J_m &\propto 1/r_g
    \label{eq:mit_kt_j} \\
    K_M^2 &\propto r_g^3 l
    \label{eq:mit_km} \\
    J_m &\propto r_g^3 l
    \label{eq:mit_j}
\end{align}
where $l$ is the axial length of the motor. 
The reference assumes constant current for thermal similarity, so we have substituted the $\tau$ used in \cite{impactmitigation} with $K_T$ without a change in meaning.
Though (\ref{eq:mit_kt_j}) appears to suggest a tradeoff in inertia, 
actuator inertia $J_a$ (\ref{eq:act_Jm}) is invariant when considering the need for a transmission to meet some given torque requirement \cite{minitaur}.

A scaling analysis in \cite{de_task_based_bldc} bridges torque production and electrical responsiveness (\textit{i.e.}, electrical time constant) as another approach, but also rewards high $r_g$ 
Thermal specific torque (\ref{eq:therm_specific_torque}) increases linearly with $r_g$ in real motors as well \cite{minitaur}.
These widely-used principles suggest that maximizing $r_g$ is a favorable design choice: it improves torque density and motor constant, thereby reducing the necessary transmission ratio and simplifying mechanical design.


\section{Proposed Metrics for QDD Motor Selection}
\label{sec:selection}
We propose alternative metrics for motor selection. Motor properties, when combined with a transmission selection, yield effective actuator performance characteristics. 
The effective torque constant, $K_{Ta}$ in [Nm/A], the effective motor constant, $K_{Ma}$ in [Nm/W$^{1/2}$], and actuator inertia, $J_a$, are defined as follows:
\begin{equation}
    K_{Ta} \doteq N\times K_T
    \label{eq:act_kt} 
\end{equation}
\begin{equation}
    K_{Ma} \doteq N\times K_M
    \label{eq:act_km}
\end{equation}
\begin{equation}
    J_a \doteq N^2 J_m
    \label{eq:act_Jm}
\end{equation}
with the transmission ratio, $N$, defined as a speed reduction.
These equations demonstrate that there is no straightforward motor and transmission ($N$) choice that balances the tradeoff between $K_{Ta}$ and $K_{Ma}$ (to be maximized) on one hand and $J_a$ (to be minimized) on the other.

To understand the relevant design trade-offs of motor selection quantitatively, we can use the following metrics, which conveniently remove the transmission selection from consideration (assuming massless transmissions with no frictional losses). 
These metrics are one way to understand the intrinsic advantages of a motor:
\begin{equation}
    \msym_M \doteq \frac{J_a}{K_{Ma}^2} = \frac{N^2 J_m}{N^2 K_M^2} = \frac{J_m}{K_M^2} 
    \label{eq:scaling_J_KM}
\end{equation}
\begin{equation}
    \msym_T \doteq \frac{J_a}{K_{Ta}^2} = \frac{N^2 J_m}{N^2 K_T^2} = \frac{J_m}{K_T^2} 
    \label{eq:scaling_J_KT}
\end{equation}
Motors that minimize $S_M$ and $S_T$ have an intrinsically better balance of properties for the characteristics of interest (maximizing $K_T$ and $K_M$ while minimizing $J_m$). 

Equation (\ref{eq:scaling_J_KM}) also has a physical implication: it approximates the mechanical time constant of motor speed response to voltage change, $\tau_\omega$, which should be minimized to improve responsiveness:
\begin{align}
    T_m &= J_m \dot{\omega} = \frac{K_T}{R_\phi} (V_q - K_B \omega) \\
    J_m \Omega(s) s &= \frac{K_T}{R_\phi} (V_q(s) - K_B \Omega(s)) \\
    \Omega(s)/V_q(s) &= \frac{ \frac{1}{K_B}}{\frac{R_\phi J_m}{K_T K_B} s + 1} = \frac{\frac{1}{K_B}}{ \frac{J_m}{K_M^2} s + 1} \\
    \rightarrow \tau_\omega &= \frac{J_m}{K_M^2} = \msym_M
    \label{eq:time_const}
\end{align}
where $T_m$ is the motor torque [Nm], $\omega$ is the motor velocity ($\Omega(s)$ in Laplace domain) [rad/s], $R_\phi$ is the motor phase resistance [$\Omega$], $V_q$ is the motor quadrature-axis (``q-axis'', \cite{mevey2009sensorless}) voltage ($V_q(s)$ in Laplace domain) [V], $K_B$ is the motor back-EMF constant [Vs/rad], and $s$ is the Laplace variable.
By (\ref{eq:time_const}), we call $\msym_M$ the Responsiveness Metric.
This analysis does not explicitly consider gap radius.
Equations (\ref{eq:mit_km}) and (\ref{eq:mit_j}) suggest that responsiveness $\msym_M$ should be invariant to choices in motor geometry under the assumptions of common ideal motor models.
Therefore, this metric is particularly useful at identifying motors that gain performance benefits from unmodeled factors (e.g., winding quality, stator core material, \textit{etc.}).
Metric $\msym_T$, which we call Torque-Specific Inertia, does not have such an invariance, but is nonetheless relevant to actuator implementation.

As established in the previous section, robot performance is also sensitive to motor mass.
Therefore, a designer may choose to incorporate a mass weighting (\textit{e.g.}, using $m\msym_M$ or $m\msym_T$).

 \subsection{Case Study: T-Motor RI50 vs. U8}
 \label{sec:casestudy}
Considering the many motor selection methodologies that emphasize large gap radius (Section \ref{sec:principles}), it is no surprise that one of the most commonly used motors in legged robots is the large radius, exterior-rotating T-Motor U8 and ``8108'' size equivalents (\textit{e.g.}, iFlight EX8, SunnySky M8) \cite{minicheetah, minitaur, pennjerboa, azocar2020design, arm_spacebok, li2021design_flying_humanoid, nesler_orthoses_2022, duperret2016core} (U8 properties in Table \ref{tab:motordata}). 
For these robots, the U8 style motor is often the largest, commonly available motor that will meet design size requirements.
The U8 is also used as the jumping-off point and benchmark to propose new motors in \cite{de_task_based_bldc}.
Due to its ubiquity, we use the U8 as a benchmark.

In contrast, the interior-rotating T-Motor RI50 is a motor that has a better balance between torque constant, motor constant, and inertia as measured by the proposed metrics (\textit{i.e.}, low $\msym_M$ and $\msym_T$, see Table \ref{tab:motordata}) than the dimensional scaling analysis would suggest for its small gap-radius \cite{impactmitigation}, and is only 193~g.

\section{Empirical Characterization of RI50 Motor}
\label{sec:characterization}
To facilitate direct comparison, we follow the U8 characterization methods described in \cite{lee_empirical} to characterize the RI50 motor.
We constructed a dynamometer that features two actuators (motors) in opposition with outputs connected through a rotary torque sensor (TRD605-18Nm, TRS605-5Nm, Futek, Irvine, CA).
Both actuators may act as the loading or driving actuator, and can swap roles mid-test as necessary.
The motor drivers (moteus r4.5/r4.8, mjbots, Cambridge, MA, USA) can be queried for position, velocity, q-axis current, and diagnostics.
A host Linux single-board computer (4B, Raspberry Pi Foundation, Cambridge, UK) commands and queries all devices and logs data using custom software written in C++.

\subsection{Passive Properties}
\label{subsec:methodpassive}

While passing current between two phase terminals (``line-to-line'') to heat the windings, we observed the motor with a thermal camera (A655, Teledyne FLIR, Wilsonville, OR, USA) to visually determine the winding style (``Wye'' or ``Delta'').

Line-to-line resistance was measured with a multimeter (34461A, Keysight, Santa Rosa, CA, USA), and line-to-line inductance was measured with a precision LCR meter (TMPRO 4230, Wayne Kerr, West Sussex, UK).

For both resistance and inductance, the value between all three combinations of phase terminals were measured and then averaged to yield a final result, which was then converted to phase-frame (phase resistance $R_\phi$ and effective inductance $L_e$) with knowledge of the winding style, per \cite{mevey2009sensorless}.

\subsection{Motor Coefficients}
\label{subsec:methodconstants}
To empirically determine the torque constant, $K_T$, for the RI50 motor, the motor under test was commanded with q-axis current using the embedded field-oriented controller (FOC) for 5~s while mechanically stalled. 
For each $i_q$ command 5~s test (-8 to 8~A in 1~A increments), the first and last 0.5~s of torque data were discarded to remove any transient artifacts.

The back electromotive force (back EMF, or BEMF) constant, $K_B$ [Vs/rad], describes the proportionality between rotational speed and the q-axis voltage ($V_q$).
To empirically determine $K_B$, the motor under test was back-driven by a cordless drill while the phase line-to-line voltages were measured with an oscilloscope.

The fundamental motor constant, $K_M$ can then simply be calculated as $K_M= \sqrt{K_T K_B/R_{\phi}}$

\subsection{Motor Inertia}
We determined the motor dynamic properties via random-input testing to observe the frequency response of the system, as in \cite{lee_empirical}. 
The motor under test was disconnected from the motor driver such that the phase leads were open circuits.
A driving actuator was commanded with a Gaussian random process (i.e., white noise) low-pass filtered to 40~Hz input and scaled to a current command in the ranges [-0.5,0.5]/$K_{Ta}$ and [-1.0,1.0]/$K_{Ta}$.
Torque data were collected from the torque transducer and velocity data were collected from the actuator encoder (sampled at 40kHz and filtered onboard the motor driver) together at 1100~Hz.

These data were fit with a non-parametric model in the frequency domain via Welch’s method in the Matlab \texttt{tfestimate()} function \cite{welch_use_1967}.
Then, a first-order parametric model was fit to the non-parametric estimate by optimizing a fit to the gain frequency response, then finally optimizing for time-domain Variance Accounted For (VAF) to obtain the final parameter estimate, both with \texttt{fmincon()}.
We chose a time-domain metric (VAF) for the second optimization stage as fitting in the frequency domain overemphasizes frequencies with low signal energy contribution.

For this test, the rotor was press-fit and bonded to a plastic arbor. 
Because the motor inertia $J_m$ was very low compared to the support hardware (shaft couplers, bearings, etc.), $J_m$ was estimated by (1) conducting the test with the rotor installed behind a lightweight $N=7.5$ planetary transmission to yield inertia $J_r$, then (2) repeating the test without the rotor and arbor to yield $J_{nr}$. 
The motor inertia $J_m$ was then taken as $J_m = \frac{1}{N^2}(J_r - J_{nr})$.

The RI50 is frameless, so its rotor mass was independently measured. 
As an alternative to the random-input test result, $J_m$ was estimated via thin-ring approximation with size and mass measurements of the rotor.

\section{Results}
\label{sec:results}

The results (summarized in Table \ref{tab:motordata}) indicate that the RI50 motor is more suitable for use in QDDs than the T-Motor U8, according to the design objectives established in Section \ref{sec:principles}.
For the RI50, a lower $K_T$ and $K_M$ can be compensated for by a higher gear ratio (\ref{eq:act_kt}, \ref{eq:act_km}), while still maintaining a lower reflected output inertia (\ref{eq:act_Jm}), compared to an actuator with the U8 or equivalent motor.
In other words, for transmission selections that result in matching actuator inertias ($J_a$), the RI50 actuator will produce more torque per current input and per heat waste than the U8 actuator (higher $K_{Ta}$ and $K_{Ma}$).

It is particularly important to note that $\msym_M$, which should be invariant to motor geometry under commonly used models (\ref{eq:time_const}), indicates that the RI50 motor is $\sim 3.4 \times$ more responsive than the U8.
This benefit is visible in clearer terms and in context with more motors in Fig. \ref{fig:ideal}.
While the predicted trend of $J_m \propto K_M^2$ (\ref{eq:mit_j}, \ref{eq:mit_km}) is visible, there may be almost an order of magnitude variation of $\msym_M$ (traversing between isolines in the figure) for some given $K_M$ range.
These variations offer designers an opportunity for higher performance. 
It also appears that TQ RoboDrive motors are particularly well suited for QDDs over a broad range \cite{noauthor_robodrive_nodate}.
Physical differences between the motors and potential drawbacks of the RI50 are discussed in Section \ref{sec:discussion}.

\begin{table}[h]
    \setlength\tabcolsep{3pt}
    \caption{Case study motors. (T-Motor) RI50 characterized in this paper and (T-Motor) U8 characterized in \cite{lee_empirical}.}
    {\centering
    \begin{center}
    \begin{tabular}{rcl|ccc}
     & $  $  &  & RI50 & U8  \\ \hline \hline
Winding Style & $  $  &  & Wye & Delta \\
Phase Resistance & $ R_\phi $  & [$\Omega$] & 0.705 & 0.279 \\
Effective Inductance & $ L_e $  & [{\textmu}{H}] & 2.559 & 0.069 \\
Torque Constant & $ K_T $  & [Nm/A] & 0.105 & 0.14 \\
Motor Constant & $ K_M $  & [Nm/W$^{0.5}$] & 0.12 & 0.23 \\
Motor Inertia & $ J_m $  & [gcm$^2$] & 90.1 & 1200 \\
Mass & $ m $  & [g] & 193 & 242* \\
Gap Radius & $ r_g $ & [mm] & 14.6 & 40.8 \\
Length & $ l $ & [mm] & 16.0 & 8.0 \\ \hline \hline
Responsiveness Metric (\ref{eq:scaling_J_KM}) & $ \msym_M $  & [ms] & 0.67 & 2.3 \\
Torque-Specific Inertia (\ref{eq:scaling_J_KT})  & $ \msym_T $  & [g(A/N)$^2$] & 0.82 & 6.1 \\ \hline
    \label{tab:motordata}
    \end{tabular}
    \end{center}
    }
    
    *The U8 is a framed motor (assembly with bearings and support structure) while the RI50 is frameless. A hypothetical frameless U8 may be $\sim$25\% lighter, but these are not yet commercially available.
\end{table}

\subsection{Passive Properties}
\label{subsubsec:resultpassive}

\begin{center}
    \begin{figure}[]
    \centering
    \includegraphics[width=0.7\columnwidth]{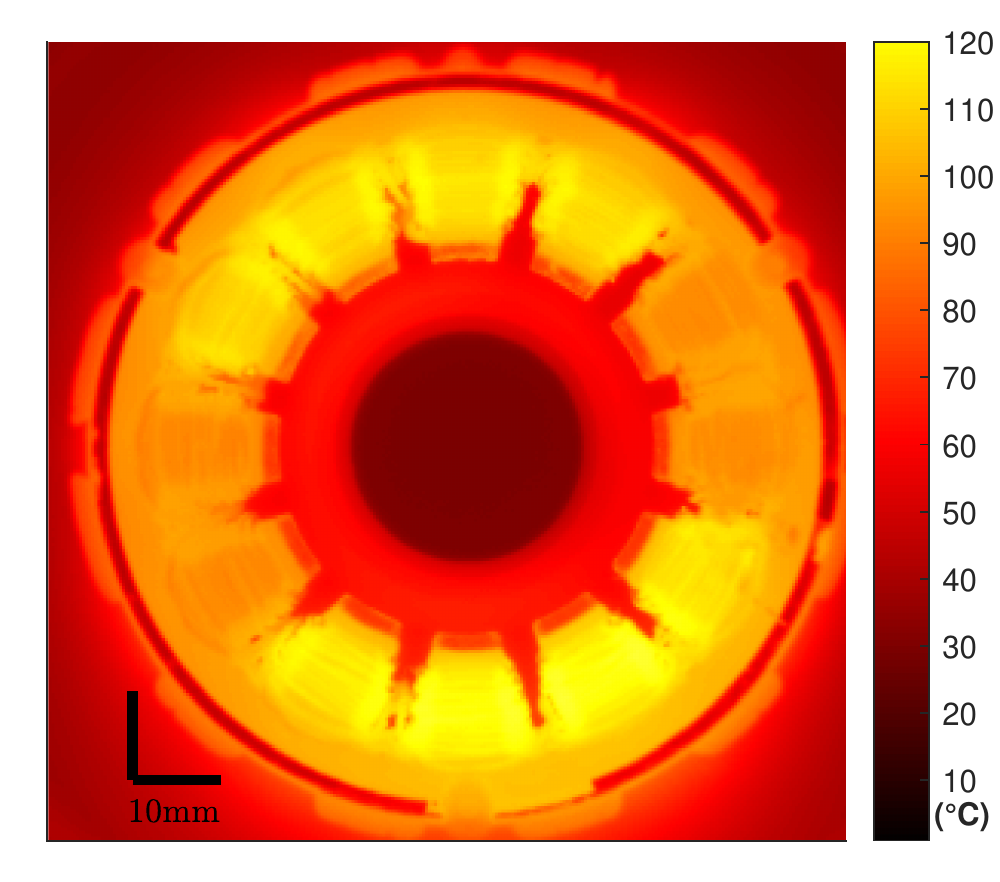}
    \caption{Averaged thermal image over 50 frames of motor during thermal test. Eight of the twelve windings appear hotter than the rest, indicating a Wye winding style.}
    \label{fig:thermal_bw}
\end{figure}
\end{center}

Thermal tests revealed that the RI50 motor is wound in the Wye style by the pattern of eight hot windings to four cold (Fig. \ref{fig:thermal_bw}).
Electrical measurements yielded a phase resistance of $R_\phi = 705\pm 3$ m$\Omega$ and an effective inductance of $L_e = 2559\pm 2$ {\textmu}H.

\subsection{Motor Coefficients}
\label{subsubsec:resultconstants}
The gradient of a line fit to the torque-current data yielded a torque constant of $K_T = 0.105\pm 0.002$ Nm/A.
The gradient of a line fit to the voltage-speed data yielded a BEMF constant of $K_B = 0.094\pm 0.002$ Vs/rad. 
Deviation from ideal PMSM patterns ($K_T = K_B$ \cite{mevey2009sensorless}) may be due to imperfectly sinusoidal back-EMF profiles.
With these values for $K_T$ and $K_B$, we can then obtain the value of the fundamental motor constant ($K_M$) as 0.118$\pm 0.003$ Nm/W$^{0.5}$.

\subsection{Motor Inertia}
\label{subsubsec:resultdynamic}
\begin{figure}[]
    \centering
    \includegraphics[width=0.95\columnwidth]{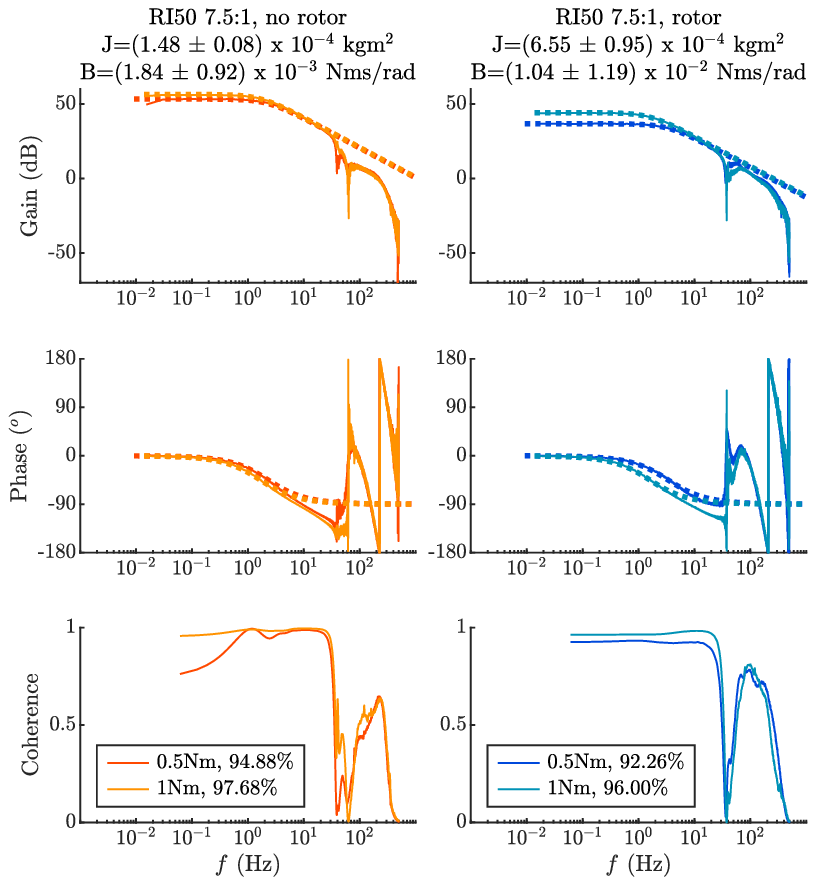}
    \caption{Mechanical System ID Frequency Response and Coherence for the 7.5:1 with and without the rotor. Legend indicated input amplitude and Variance Account For goodness of fit. The confidence intervals for the $J$ and $B$ are 2 standard deviations of identified $J$ and $B$ across input magnitudes}
    \label{fig:mech_sysid}
\end{figure}

The frequency domain non-parametric estimate had high coherence up to 40~Hz (Fig. \ref{fig:mech_sysid}), suggesting that the torque and velocity data could be explained by a first-order linear model:
\begin{equation}
    \frac{\Omega (s)}{ T_b (s)} = \frac{1}{J_e s + B_e}
    \label{eq:mech_dyn_model}
\end{equation}
where $\Omega(s)$ is actuator speed, $T_b (s)$ is backdriving torque, $J_e$ is output inertia [kgm$^2$], $B_e$ is viscous damping [Nms/rad], and $s$ is the Laplace variable, and where the subscript $e \in \{r, nr\}$ for rotor installed and not installed, respectively. 
The following parameters resulted from fitting the model: the actuator inertia $J_r$ was estimated as $(6.55 \pm 0.95) \times 10^{-4}$ kgm$^{2}$ and $J_{nr}$ was estimated as $(1.48 \pm 0.08) \times 10^{-4}$ kgm$^{2}$.
The goodness of fit is evaluated with the Variance Account For (VAF) metric.
The parametric fit showed $92\%-98\%$ VAF, indicating a high-quality parameter estimation for the two inertias.
These two results yielded a final inertia estimate of $J_m = (9.01 \pm 1.8) \times 10^{-6}$ kgm$^{2}$.

The thin-ring inertia estimate yielded $J_m$ as $(5.4 \pm 0.5) \times 10^{-6}$~kgm$^2$ with a mass of $34\pm 0.5$~g and radius of $12.6\pm 0.5$~mm.
Differences between the random-input result and thin-ring result are likely due to missing inertial contributions from support bearing inner-races.
These contributions are not captured with the rotor (and arbor) uninstalled.

\section{Discussion}
\label{sec:discussion}

We propose two motor selection metrics, $S_M$ and $S_T$, for use in Quasi-Direct Drive actuator design.
These metrics are simple, invariant to transmission ratio selection, and readily evaluated on a wide range of motors via datasheet values (Fig. \ref{fig:ideal}).
While maximizing gap radius $r_g$ has been a successful strategy in QDD actuator design (Section \ref{sec:principles}), we show through a case study in comparing the T-Motor U8 and RI50 that alternative motors may be more suitable, as illuminated by the proposed metrics.
The mechanical and electrical properties of the RI50 motor is rigorously characterized (Section \ref{sec:results}) for accurate comparison to the U8 motor.
Designers may leverage the proposed metrics to (1) quickly down-select from a range of motors, and (2) find motors that offer different, potentially useful, design opportunities.
For example, the RI50 has comparable performance properties to the U8, but is much more compact (Fig. \ref{fig:motorpics}), offering alternative packaging strategies than with the U8.

The RI50's performance characteristics are indeed surprising, given its very small gap radius.
Here, we propose explanations for the unexpected performance advantages of the RI50.

\subsection{Coil Turns Contribute to $K_T$}
Common motor analyses focus on geometric parameters and use assumptions to remove number of coil turns, $n_t$, from consideration \cite{impactmitigation, seok_actuator_2012}.
However, $n_t$ is a useful ``lever'' for motor designers to use in manipulating motor properties \cite{hanselman2003brushless}.
The parameter $n_t$ is linearly related to the shear force generated between the stator and rotor in the motor, as it amplifies the magnetic field of the motor coil.
Thus, $K_T \propto n_t$.
In the case of the U8 motor, $n_t \approx 16$ and for the RI50, $n_t \approx 36$ by visual inspection.
This greater $n_t$ allows the RI50 motor to partially make up for its radius disadvantage with respect to $K_T$, and also explains the higher $R_\phi$ and $L_e$ for the RI50.
This effect is captured by the $\msym_T$ metric (\ref{eq:scaling_J_KT}), and is also discussed in \cite{de_task_based_bldc}.

\subsection{Inner- vs. Outer-rotating Geometry Greatly Affects $K_M$}
The RI50 is inner-rotating (its rotor is inside its stator) and the U8 is outer-rotating. 
With the small gap radius of the RI50, this difference plays an important role that common geometric analyses do not capture.
Specifically, it is often assumed that the rotor with radial thickness $t_r$ and stator with radial thickness $t_s$ (Fig. \ref{fig:motorpics}) are thin rings, \textit{i.e.}, $t_r \ll r_g$, $t_s\ll r_g$ \cite{impactmitigation}.
While this assumption is approximately true for the rotor, it is not for the stator in the case of the RI50.
This fact means that the stator can hold more copper, 
lending it a higher $K_M$
than an outer-rotating motor of the same gap radius and length \cite{hanselman2003brushless}.
Such an outer-rotating motor would have $\sim 4 \times$ less volume available for the stator.
This effect is captured by $\msym_M$ metric (\ref{eq:scaling_J_KM}).

\subsection{Additional Design Considerations}
The proposed motor selection strategy emphasizes properties useful for QDD actuator design.
However, additional properties can be considered for specialized applications.
For example, although the RI50 has high torque for its inertia and is lighter than the U8, its torque density, $K_T/m$, is lower than the U8 --- designers should determine whether the higher transmission ratio necessary to meet torque requirements adds untenable mass or complexity.
The RI50 also has a very high inductance, $L_e$ thereby limiting current-control bandwidth, which could potentially limit performance if the bandwidth is not sufficiently faster than robot mechanical dynamics.
Lastly, the RI50 has a high phase resistance, which will naturally limit its peak speed.
Higher voltages can be used to compensate, but may be limited by other system components (\textit{e.g.}, the motor driver).
Future work will consider the relative importance of these drawbacks in the context of a full robot.%

\bibliographystyle{IEEEtran}
\bibliography{main.bib}

\newpage

\vfill

\end{document}